# Stealth UAV through Coandă Effect

Dongyoon Shin
Computer Science and Engineering
Sogang University
Seoul, Republic of Korea
ansel93@sogang.ac.kr

Hyeji Kim
Computer Science and Engineering
Dongguk University
Seoul, Republic of Korea
hyex77@dongguk.edu

Jihyuk Gong
Computer Engineering
Jeju National University
Jeju, Republic of Korea
kongji4092@gmail.com

Uijeong Jeong
Computer Science and Engineering
Dongguk University
Seoul, Republic of Korea
mjsk0583@dongguk.edu

Yeeun Jo
Computer Science and Engineering
Ewha Womans University
Seoul, Republic of Korea
grace_jo@ewhain.net

Eric Matson
Computer and Information Technology
Purdue University
West Lafayette, United States of America
ematson@purdue.edu

*Abstract*— **This paper uses Coandă Effect to reduce motors, the source of noise, and finds low noise materials with sufficient lift force so that it can achieve acoustical stealth UAVs.According to NASA research [1], the noise of UAVs is better heard to people. But there must be some moments when we need to operate the drones quietly, so how can we reduce the noise? In previous research, there have also been steady attempts to produce UAVs using Coandă Effect, but have never tried to achieve Acoustic Stealth through Coanda UAVs. But Coandă Effect uses only one motor and is structurally quiet. So we tried to find quiet methods (materials, structures) while at the same time having sufficient stimulus through the Coandă Effect. Verification went through experiments. The control group used the most common type of Quadrone, and determine if the hypothesis is correct by testing various structures and materials under the same conditions, and measuring noise. UAVs using Coandă Effect are not of any shape or structure that is not changeable, and internal space is also empty. That's why the Coandă Effect UAV we present can be improved through follow-up research. That's why the Coandă Effect UAV could open up a new frontier for the Stealth UAVs.**

## I. Introduction

UAV(Unmanned Aerial Vehicle), which was initially developed for military purposes, is now commercialized and used in more than half of the world's countries to be widespread among international organizations, governments, businesses, academia and individuals. Flying UAV are very attractive military equipment. It is cheaper than other equipment, easier to control, and doesn't ride person.

Another reason why UAV have been spotlighted in the military arena is that the use of UAV is the casualty of their soldiers. It is said that while reducing the number, only the target person can attack, which reduces the risk considerably.[2] Since UAV are used as military equipment, stealth functions are required to prevent them from being detected by the enemy.

Coandă UAV had already used as a military UAV in USA, Russia and UK, but they stopped developing Coandă UAV because of technical reasons. Unlike past, small size UAVs are common for military use and more effective for stealth. For big size UAV because of Coandă UAV may have not enough lift force and ineffective, but they are better for small stealth UAV.

There are four main ways to detect UAV. Typical methods are radar, search through noise, checking engine heat and direct eye check, using electromagnetic waves. Stealth function on decibels were intensively studied in the paper.

In this paper, we proposed a method to reduce the number of motors in order to reduce the sound of motors that produce the most noise among components, absorb the sound of motors, and apply a cloth to the surface of UAV to reduce noise.

UAV has a different name depending on the number of motors. Quadcopter UAV with four motors take the shape of UAV that are most commonly known is the noise of the motor that makes up a significant part of the noise of the drone. However, if the number of motors is reduced to one to reduce noise, the power to support motors will be weaker than that of quadcopter. So we proposed using the physical effect of the Coandă Effect as a solution to support the UAV, and as a result, we made a UAV using only one motor. However, using only one motor has limited weight to support. Therefore, the frame of the UAV was built using a 3D printer that could make the design fluid. Moreover, it can make frames that are fairly light compared to metals. Our UAV reduces the number of motors and implements silence-free drones through cloth-based absorption and reduces noise compared to commonly used quadcopter UAV.

## II. STEALTH UAV

### A. Coandă Effect

The Coandă Effect is an effect discovered in 1930s by Henri Marie-Coandă. His patent explains this effect as "If at high speed one gas is released into the atmosphere of another of some kind, if the gas is inhaled and pulled forward of the adjacent gas, this will be generated at the release point of that gas.""[3] "On the outside of a fluid flow or sheet, if an imbalance effect is set on the flow of the surrounding fluid induced by that fluid flow, the latter will move to the direction where the flow of the surrounding fluid has become more difficult [3]." Which to be concrete, the tendency of a fluid jet to stay attached to a convex surface. The theoretical background is as follows.

"We are considering that the opening slit width of the section is $b_0$, it has its normal line perpendicular on the jet axes (Figure 1) and because of the presence of the Coandă profile on one of its sides that limit's it an asymmetrical flow appears (Figure 2) which remarks itself through an asymmetrical distribution of the gasodynamic parameters in that section.

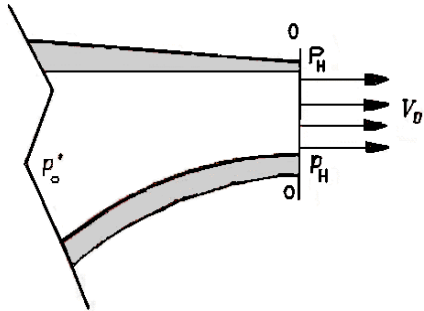

Figure 1. Slit flow of free jet

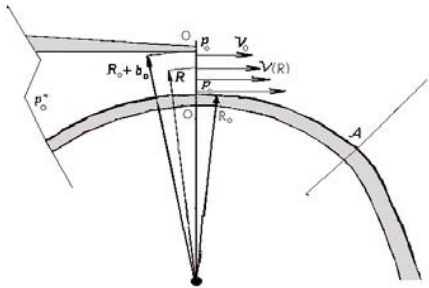

Figure 2. Flow with Coandă Effect

The distribution of the speed of slackening V0 from $p_0^*$ to the ambient pressure pH is:

$$V_0 = \sqrt{2(i_0^* - i_H)} = \sqrt{2i_H^* \left[\left(\frac{p_0^*}{p_H}\right)^{\frac{k-1}{k}} - 1\right]} \quad (1)$$

The variation of speed depending on the radius, it is thought to be:

$$V(R) = V_0 \left(\frac{R_0 + b_0}{R}\right)^n, \text{ and we have}$$

$$V_0 = V(R_0 + b_0) \quad (2)$$

In the case were the radius of the slit curb is bigger compare to the slit opening, we can make the approximation that the brake enthalpy is constant on the radius and the asymmetrical effect is caused by the variation of the static pressure from pH in the upper part of the slit, till p0 on the wall in the lower part of the slit."[4]

The Coandă Effect is used in most aircrafts to generate sufficient lift for support. As the plane moves rapidly, the air splits off by wing. Wing of aircraft, which is so called airfoil shaped, made air that rises to the upper side of the aircraft's wing descends downward on wings by the Coandă Effect, while air that goes lower side of the wing remains. As a result, the air gathered at the bottom. This result makes higher air pressure at the bottom of the aircraft than at the top of the aircraft, creating in a greater lift than the mass of the aircraft.

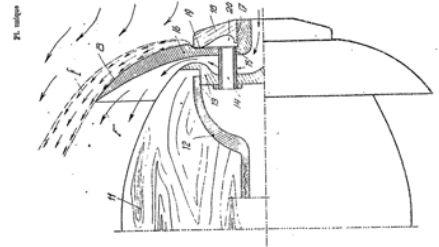

Figure 3. Coandă patent "Perfectionnement aux propulseurs"[5]

Even if it's not a typical airplane, it's possible to build a UAV which focused on the Coandă Effect, Henri Marie-Coandă, who discovered the Coandă Effect have already designed s form that can be supported using the Coandă Effect in 1935(Figure3). In addition, when referring to the various recently developed Coandă Effects UAVs [6], we determined that using just one propeller would give us enough lift force.[7]

The noise of drones is mostly based on motors and propellers.[8] This noise can be classified into the noise of the propeller itself and the noise caused by the vibration produced during the propeller as it delivered air.

In the case of propeller-induced noise, the motor and propeller will be limited to one by designing of the Coandă Effect UAV [6], which naturally reduced the volume. However, we don't only reduce the number of propeller but also try to increase its effectiveness of noise reduction by using a longer diameter propeller.[9]

For vibration issue, multi-copters, such as Quadrones or Hexadrones, vibration amplification due to multiple motors can result in increased noise, but vibration can be greatly reduced if only one propeller is used by using Coandă Effect.

We propose several ways to reduce vibration in the relevant study,[10] where we came up with the Velcro as a connection between the control system and the drone body. This can reduce vibrations of drone body and reduce overall noise.

In addition, while we are using the shape that take advantage of Coandă Effect, development of UAV is autonomous in all aspects and able to develop in any form. [11] UAVs in our paper can be used to create improved stealth drones by applying additional existing or developed stealth technologies. It is not a finished stealth UAV, but can be seen as a basic form of a drone that is advantageous for acoustic stealth function.

*B. 3D Modeling*

Manufacturers and product developers have used additive manufacturing, a process more commonly known as 3D printing, to create prototypes, mock-ups, and replacement parts.[12] 3D printing is a disruptive technology that promises to change the way we consume, create, and maybe even live in the world. Some are even calling 3D printing and attendant digitization of manufacturing the third industrial revolution. [13]

3D printing offers the promise of control over the physical world. In a 3D printed future world, people will make what they need. To design a UAV that has a low profile, the three-dimensional objects were designed and printed using the cutting-edge technique, 3D printing. 3D printer fabricates three-dimensional objects when it is fed a well-designed electronic blueprint, or design file, that tells it where to place the raw material. [14]

Before printing, we designed 3D models of UAV frames with Autodesk® Fusion 360. After modeling, we switched each 3D model files into a form that is adequate for 3D printer. We used the Cura Lulzbot edition for this format modification process. Cura Lulzbot Taz 6, a widely used 3D printer, was used for printing the UAV frame and PLA(Verbatim) was used as the 3D printing filament.

During the Coandă Effect UAV frame designing process, we solved several robotics issues. First issue was how to divide the UAV frame separately and how to design the shape of each frame shape to make it suitable for Coandă Effect. These questions were solved by referring to previous researches. We created the Coandă Effect UAV based on Jean-Louis Naudin's GFS-UAV model N-01. The design of the GFS-UAV N-01A was based on the Geoff Hatton flying saucer which was promoted by GFS Project limited. As shown in Fig #, the Geoff Hatton's flying saucer had an octagonal shape canopy, with flat flaps on four opposite sides of the trailing edge. Later, Naudin published a paper introducing the full plan for building the GFS-UAV N-01A. [15]

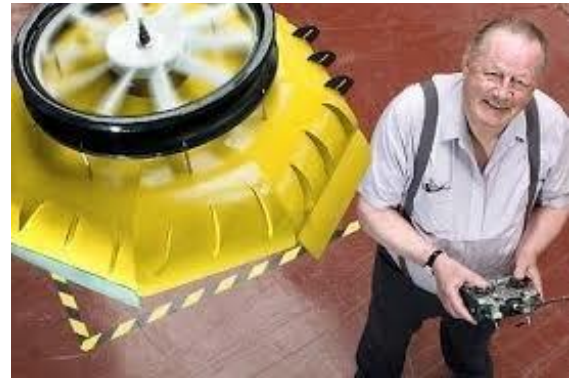

Figure 4.  Geoff Hatton and his flying saucer.[15]

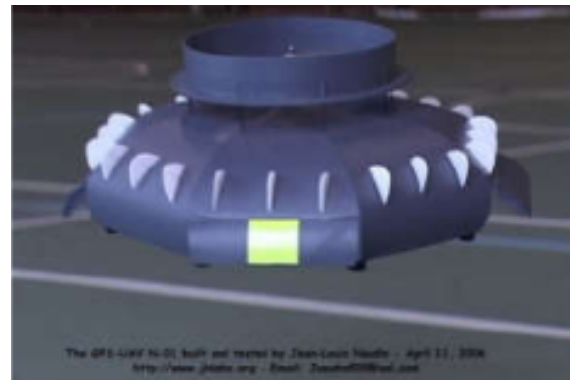

Figure 5.  J.L-. Naudin's first GFS-UAV (N-01A)[15]

The Second issue was how to combine each UAV pats into well-balanced frame. We hollowed out the 3D model design in a way that will let the particles combine like LEGO pieces. The LEGO-like particles were suitable to connect frame evenly and stable. The Third issue was figuring out the proper thickness of the frame and the proper size of the frame. It was solved by adjusting the size of the frame according to the measured size of the propeller, motor, raspberry pi and other components that will be attached inside of the frame. It was useful to print out the design and adjust the size according to each connecting trial. Given this, the consequential Coandă Effect UAV frame used in this paper is made using the particles as shown in the following figures.

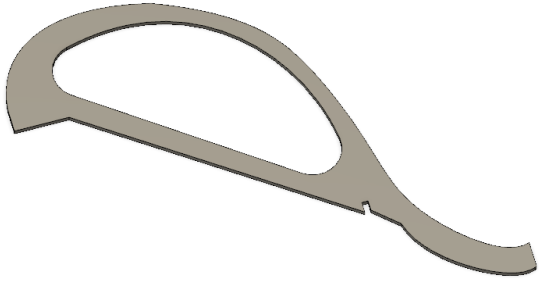

Figure 6.  Coandă Effect UAV frame : side part

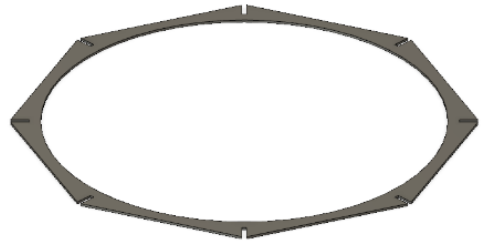

Figure 9.  Coandă Effect UAV frame : bottom part

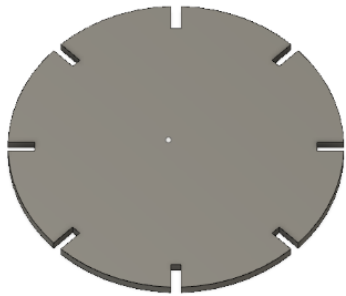

Figure 7.  Coandă Effect UAV frame : top circle

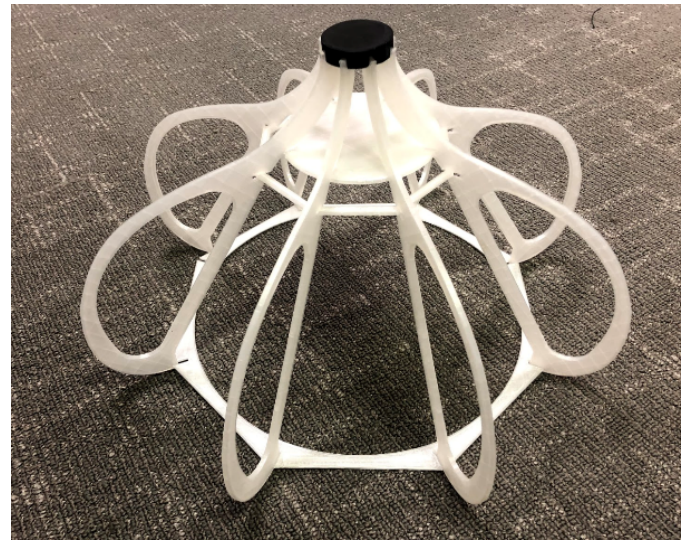

Figure 10.  Coandă Effect UAV frame all combined

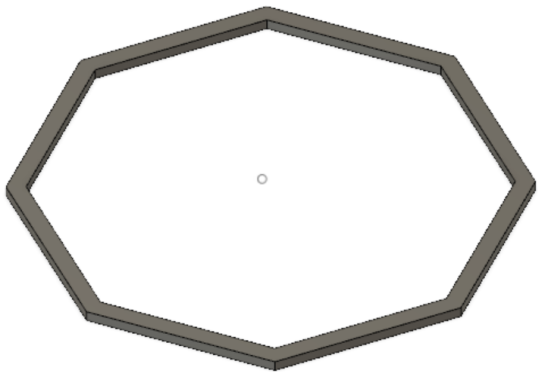

Figure 8.  Coandă Effect UAV frame : top part

There are three reasons we used 3D modeling and 3D printing method in this project.

First, 3D printing's unique manufacturing technique enables making objects in shape that were never possible before[10]. In this paper, as the parts for building a Coandă Effect UAV is a unique shape, the frame was rarely found in the market. So, using 3D modeling software to design all the unique particles of the UAV frame was appropriate.

Second, it is easy to modify a design when using 3D modeling software. Several robotics issues arouse while conducting the UAV flight simulation, which was executed in order to build the best stealth function state drone. It was convenient to solve these issues by redesigning the frames or changing the size.

Third, using 3D printing with PLA was efficient for decreasing UAV weight. The shape of the Coandă Effect UAV is designed to use a Coandă Effect when flying. It was designed to use one motor and one propeller which were fewer than the basic quad drone which has four motors and four propellers. This feature required to minimize the weight of the UAV. Therefore, it was proper to use lightweight PLA(verbatim) instead of iron.

## C. Anti-Noise With Material

The noise usually generated by our drones is aerodynamic sound caused by wind flowing along the surface of UAV. Aerodynamic sound is one of the dominant noise sources derived from high-speed trains, airplanes, fans, in electronic equipment, etc. Noise reduction is crucial in building UAVs because noise makes UAV vulnerable to detection [9]. There are several studies that applied of material to reduce aerodynamic sound, or noise. They [16][17] succeeded in noise reduction by covering an object surface with materials. We are going to apply these studies to minimize noise.

Our study focuses on changing the fabric material on the UAV surface, meeting two conditions. First is the ability to float. In order for our drone to float, the Coandă Effect should happen. we should select the materials without disturbing this effect. Second is the degree of sound reduction. Since Stealth is the purpose of our drones, we need to reduce the noise. For this reason, we should select the materials which have a high degree of sound reduction.

There are many kinds of materials, and it is essential to select the material appropriate for our purpose. Therefore, based on these two conditions, this paper experimented with three materials: Flexible cloth, Flexible and water-repellent cloth, and thick paper.

- Flexible cloth: This flexible cloth, which is a kind of polyester and spandex, would completely cover the surface of the UAV. We think that using non-tensioned cloth would change the flow of wind on the surface of an object. It is expected that the flexibility has a positive effect on the ability to float. Also, a previous experiment[18] contains the results that Polyester fabric works on sound absorption. Based on this experiment, we judged that the cloth can reduce the sound. This is set to baseline compared with other materials.

- Flexible cloth and water-repellent cloth: A flexible cloth set as a baseline is a ventilated, or airy cloth. It was expected that the wind would penetrate cloth and flow inside the airframe. In other words, the Coandă Effect would not occur and it affects negatively the floating ability. For this reason, it was necessary to choose another fabric in which the wind did not pass through the airframe. To improve this ability, we selected the Flexible cloth with water-repellent function. It also expected reducing noise because it is cloth.

- Thick paper: This "paper" is a kind of vellum paper which is matte and thick. We call this "paper" a vellum to prevent confusion. Since the vellum is light-coated on both sides, the wind cannot penetrate it. It was also expected to occur Coandă Effect because of the flexibility and tension. That is, it seemed that the vellum would have a high level of ability to float. According to this paper[19], Paper fibres posses high fibre porosity, making them ideal to be made into sound absorbers. Thus, we chose vellum that meets the two conditions as the final material.

## III. EXPERIMENTS

### A. Experimental Models

In order to effectively amplify the Coandă Effect, three prototypes with different exterior wall materials were manufactured. Because this experiment only requires flotation of UAV in one place, We simplified the complex structure of Jean-Louis Naudin UAV.[15] For comparative analysis, Quadrone, which is a common form of UAV, was used as a comparison model.

Common hardware specifications of the experimental models are as follows:

TABLE I. HARDWARE OF THE EXPERIMENTAL MODELS

| Hardware | Model name |
|---|---|
| Main motor / Propeller | QWinOut A2212 1000KV Brushless Outrunner Motor 13T, 1045 Propeller |
| Servo motor | SG90 9g Micro Servos (2 piece) |
| Battery | Hobbypower 30a Brushless Speed Controller |
| ESC | Hobbypower 30a Brushless Speed Controller ESC |
| Flight Controller | Raspberry Pi 3 Model B |

The Coandă Effect UAV's frame is designed with a 3D printer using PLA Filament, while the Quadrone's frame is made with the DJI Flame Wheel F450 Basic Quadcopter Drone Kit.

The differences between the three prototypes are as follows:

TABLE II. EXTERIOR WALL MATERIALS OF COANDĂ-EFFECT UAV PROTOYPES

| prototype # | Exterior wall material |
|---|---|
| prototype 1 | Eurmax 6Ft Rectangular Fitted Spandex Tablecloths (Black) |
| prototype 2 | Discount Fabric Polyester Lycra /Spandex 4 way stretch Solid Slate Gray Grey LY855 |
| prototype 3 | Thick Cardstock Paper |

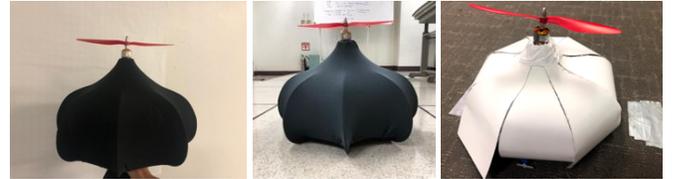

Figure 11. Prototype 1, 2, 3

### B. Experimental Environent and Methods

All experiments were performed in the same room at the same location. The background noise was measured below 40dB and found to be less than the minimum noise of the motors and propellers. UAV were connected to the ground with just enough space to hover in the midair for data measurement. We measured the noise from motors and propellers when the UAV is completely separated from the ground. The sonometer is located 24 inches from the center of the UAV, and the microphone of the sonometer is in the opposite direction of the wind to avoid the noise of the wind.

We saved the code to run the motors on the Raspberry Pi and controlled it remotely using Wi-Fi. The speed of the motor is the pulse width that the motor receives. The maximum value is 2200 and the minimum is 700. More than 270 data were collected for each experimental model, and the results of the experiments are analyzed using the means of data and the trend of all data.

*C. Results*

  *1) Quadrone*

TABLE III. NOISE OF QUADRONE

| Model | Speed | Noise (dB) | | |
|---|---|---|---|---|
| | | *Avg* | *Max* | *Min* |
| Quadrone | 1400 | 84.54 | 123 | 57 |

  *2) Prototype 1*

TABLE IV. NOISE OF PROTOTYPE 1

| Model | Speed | Noise (dB) | | |
|---|---|---|---|---|
| | | *Avg* | *Max* | *Min* |
| Prototype 1 | 1500 | 71.32 | 99 | 61 |
| | 1700 | 72.79 | 114 | 55 |

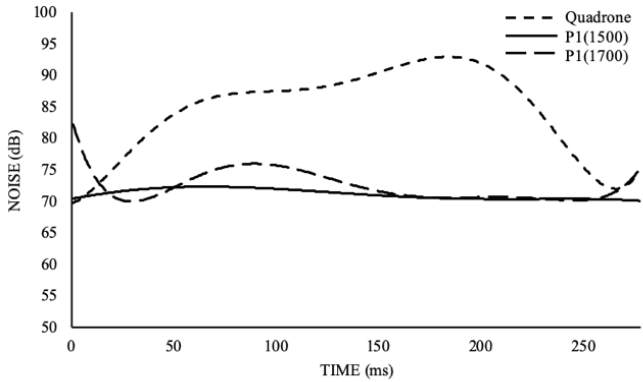

Figure 12. Trend Line graph with Quadrone and prototype 1

The minimum motor speed for Quadrone to fly was 1400 and P1(i.e.Prototype1) was 1500 but experiments were also conducted at 1700 to measure noise reduction effects. The comparison between Quadrone and P1 (speed 1500) reveals that the average value of Quadrone noise is 4.6 times higher than P1.(Table 1) Figure 2 clarifies that Quadrone's data is generally louder. The motor speed of the P1 was tested at 1700, and the Quadrone averaged 3.9 times higher. Even at high motor speeds, the Coandă Effect UAV had less noises, indicating that the Coandă Effect affects noise reduction.

  *3) Prototype 2*

TABLE V. NOISE OF PROTOTYPE 2

| Model | Speed | Noise (dB) | | |
|---|---|---|---|---|
| | | *Avg* | *Max* | *Min* |
| Prototype 2 | 1500 | 83.45 | 120 | 66 |
| | 1700 | 83.92 | 121 | 67 |

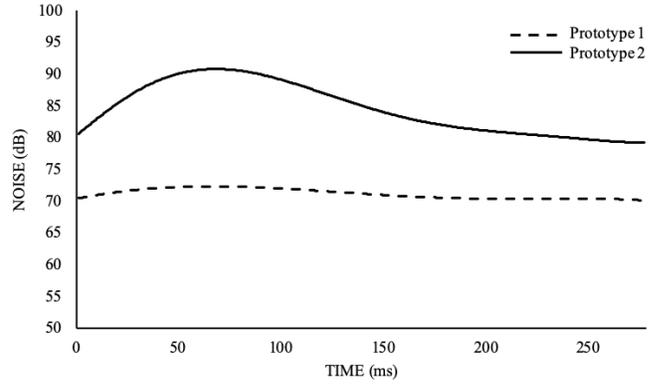

Figure 13. Trend Line graph with prototype 1 and prototype 2 at 1500

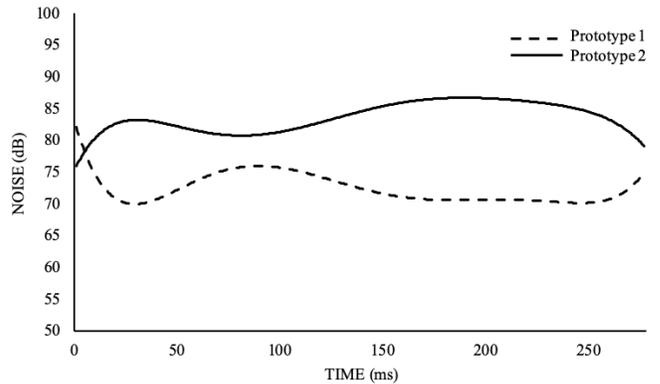

Figure 14. Trend Line graph with prototype 1 and prototype 2 at 1700

The minimum flotation speed of P2(i.e. Prototype2) is 1500 and was compared with P1 to determine if there is a reduction in noise depending on the replacement of exterior wall material. Because of the waterproof function of the fabric of P2, We expected to have a higher Coandă Effect and a smaller noise than P1. However, at the speed 1500, the P2 appeared four times the noise of the P1 and at the 1700, had 3.6 times. The P2 has a lower noise than the Quadrone, but unexpectedly, produced louder vibrations, making noise louder than the P1.

  *4) Prototype 3*

TABLE VI. NOISE OF PROTOTYPE 3

| Model | Speed | Noise (dB) | | |
|---|---|---|---|---|
| | | *Avg* | *Max* | *Min* |
| Prototype 3 | 1500 | 60.48 | 94 | 48 |
| | 1700 | 69.16 | 96 | 55 |

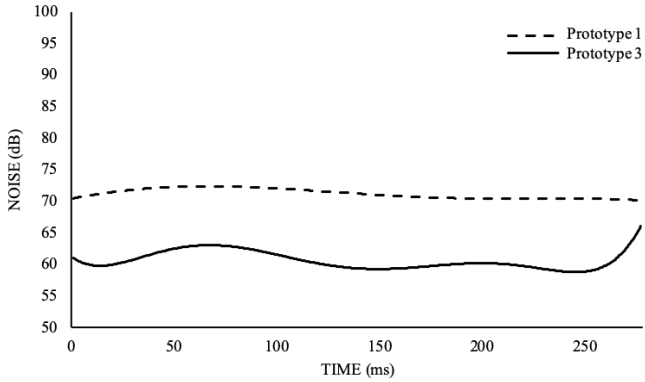

Figure 15. Trend Line graph with prototype1 and prototype 3 at 1500

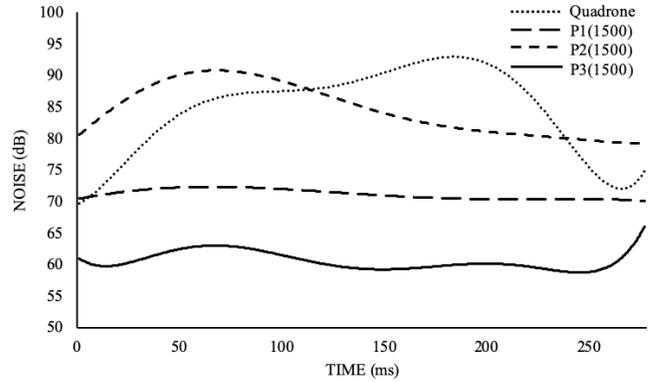

Figure 17. Trend Line graph at 1500

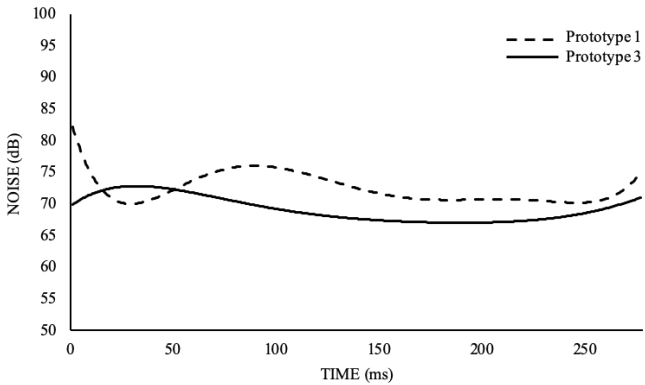

Figure 16. Trend Line graph with prototype 1 and prototype3 at 1700

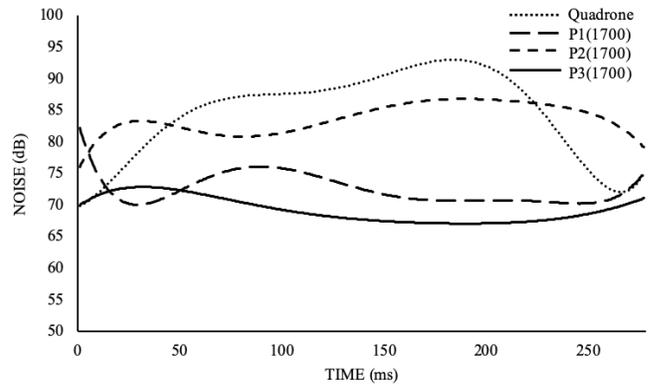

Figure 18. Trend Line graph at 1700

As a result of comparison with Prototype 1, which had the smallest noise before, it was confirmed that the noise reduction of P3(i.e. Prototype 3) was the most. At speed 1500, the average value of P1 was 3.5 times than P3 and 1.5 times at the 1700. the use of paper as the exterior wall allowed the wind to flow better than the fabric, which used in P1 and P2, emphasizing the Coandă Effect.

*5) Summary*

TABLE VII. NOISE OF WHOLE MODELS

| Model | Speed | Noise (dB) | | |
|---|---|---|---|---|
| | | *Avg* | *Max* | *Min* |
| Quadrone | 1400 | 84.54 | 123 | 57 |
| Prototype 1 | 1500 | 71.32 | 99 | 61 |
| | 1700 | 72.79 | 114 | 55 |
| Prototype 2 | 1500 | 83.45 | 120 | 66 |
| | 1700 | 83.92 | 121 | 67 |
| Prototype 3 | 1500 | 60.48 | 94 | 48 |
| | 1700 | 69.16 | 96 | 55 |

Experiments with both speeds were able to identify the tendency to have less noises in the order of Quadrone, P2, P1, P3. (fig 7, fig 8) The P3, which uses the best wind-resistant material, has the greatest noise reduction. The results clarifies that the better the Coandă Effect is applied, the greater the noise reduction effect. Compared to the P3, the average noise of the Quadrone is 16 times higher at 1500 and 5.9 times higher at 1700.

IV. CONCLUSIONS

This paper discusses the Stealth UAV using the Coandă Effect. We hypothesized that the Coandă Effect has a significant advantage in noise reduction, and focused on noise reduction by adding weight savings and anti-noise materials through 3D printing. Through the noise comparison experiment with Quadrone, the Coandă Effect UAV manufactured in this paper shows noise reduction effect of up to 16 times. The results derives that the Coandă Effect has an advantage in noise reduction when applied to the stealth UAV.

ACKNOWLEDGMENT

This research was supported by the MSIT(Ministry of Science ICT), Korea, under the National Program for Excellence

in SW (2015-0-00910) supervised by the IITP(Institute for Information& Communications Technology Planning & Evaluation)

## REFERENCES

[1] Initial Investigation into the Psychoacoustic Properties of Small Unmanned Aerial System Noise, Andrew Christian and Randolph Cabell, NASA Langley Research Center, 2017

[2] A study on the trend of anti-drone technologies and their applications, Jung, Jeyong, Chun, Yong-Tae,Korea Security Science Association, 2017

[3] Coandă, H.. "Device for Deflecting a steam of Elastic Fluid Projected into an Elastic Fluid". US Patent Office, US Patent # 2,052,869, 1936

[4] Ionică CÎRCIU, Sorin DINEA, "REVIEW OF APPLICATIONS ON COANDĂ EFFECT. HISTORY, THEORIES, NEW TRENDS", Review of the Air Force Academy The Scientific Informative Review, No 2(17), pp. 14-20, 2010.

[5] Coandă, H., Perfectionnement aux propulseurs, Brevet d'invention France, no. 796.843, 15.01.1935

[6] Florin Nedelcut, "COANDĂ EFFECT UAV – A NEW BORN BABY IN THE UNMANNED AERIAL VEHICLES FAMILY", Review of the Air Force Academy The Scientific Informative Review, No 2(17), pp. 21–28, 2010.

[7] Daniela FLORESCU, Iulian FLORESCU, Florin NEDELCUȚ, Iulian NEDELCU, "FUSELAGE AIRSTREAM SIMULATION FOR A COANDĂ UAV", Review of the Air Force Academy The Scientific Informative Review, No 2(17), pp. 83-87, 2010.

[8] Alex Stoll, DESIGN OF QUIET UAV PROPELLERS, Stanford University, 06.2012

[9] Jiyeon Oh, Daeun Choe, Chanhui Yun, and Michael Hopmeier, "Towards the Development and Realization of an Undetectable Stealth UAV", IEEE IRC, 28.03.2019

[10] Zhenming Li, Mingjie Lao, Swee King Phang, Mohamed Redhwan Abdul Hamid, Kok Zuea Tang, and Feng Lin Mechanical Engineering, National University of Singapore, Singapore Temasek Laboratories, National University of Singapore, Singapore Electrical and Computer Engineering, National University of Singapore, Singapore. Development and Design Methodology of an Anti-Vibration System on Micro-UAVs. International Micro Air Vehicle Conference and Flight Competition (IMAV), 2017.

[11] Md. Enamul Haque, Mohammad Mashud, Md. Nazmul Hasan, "Unmanned Aerial Vehicles Construction by Coandă Effect", ICERIE, 2017

[12] Peacock, Sklyer R. "Why manufacturing matters: 3D printing, computer-aided designs, and the rise of end-user patent infringement." William & Mary Law Review 55.5 (2014): 1933.

[13] J. Dale Prince (2014) 3D Printing: An Industrial Revolution, Journal of Electronic Resources in Medical Libraries, 11:1, 39-45, DOI: 10.1080/15424065.2014.877247

[14] Lipson, Hod, and Melba Kurman. Fabricated: The new world of 3D printing. John Wiley & Sons, 2013. p.11-12

[15] Haque, Md & Shafayate Hossain, Md & Assad-Uz-Zaman, Md & Mashud, Mohammad. (2015). Design and Construction of an Unmanned Aerial Vehicle Based on Coandă Effect.

[16] Takeshi Sueki, Takehisa Takaishi, Mitsuru Ikeda and Norio Arai. (2008). Application of porous material to reduce aerodynamic sound from bluff bodies

[17] Allen Powell (1964) Theory of vertex sound.

[18] Nik Normunira Mat Hassan, Anika Zafiah M. Rus. (2013). Influences of Thickness and Fabric for Sound Absorption of Biopolymer Composite

[19] Jason S.T.Sim, Rozli Zulkifli, M.F.Mat Tahir, A.K.Elwaleed. (2014). Recycled Paper Fibres as Sound Absorbing Material